\documentclass[11pt,a4paper]{article}
\usepackage[hyperref]{acl2020}
\usepackage{times}
\usepackage{latexsym}
\usepackage{booktabs}
\usepackage[margin=0.5in]{geometry}
\usepackage{graphicx}
\usepackage{subfig}
\usepackage{bm}
\usepackage{enumitem}

\usepackage{microtype}

\aclfinalcopy 

\graphicspath{{images/}}

\title{Action Quality Assessment using Transformers}

\author{Mohammad Alali \\
  \texttt{alalim@usc.edu} \\\And
  Hemanth Bodala \\
  \texttt{bodala@usc.edu} \\\And
  Abhay Iyer \\
  \texttt{adiyer@usc.edu} \\\And
  Sunit Vaidya \\
  \texttt{sunitash@usc.edu} \\
}

\date{}

\begin{document}
\maketitle

\begin{abstract}
    Action quality assessment (AQA) is an active research problem in video-based applications that is a challenging task due to the score variance per frame. Existing methods address this problem via convolutional-based approaches but suffer from its limitation of effectively capturing long-range dependencies. With the recent advancements in Transformers, we show that they are a suitable alternative to the conventional convolutional-based architectures. Specifically, can transformer-based models solve the task of AQA by effectively capturing long-range dependencies, parallelizing computation, and providing a wider receptive field for diving videos? To demonstrate the effectiveness of our proposed architectures, we conducted comprehensive experiments and achieved a competitive Spearman correlation score of 0.9317. Additionally, we explore the hyperparameters effect on the model’s performance and pave a new path for exploiting Transformers in AQA.
    
\end{abstract}

\section{Introduction}
\label{sec:introduction}

Assessing an action quality is an active field that has not been solved yet; moreover, at the core of most solutions is a convolutional operator that provides the feature extraction mechanism such as I3D + MLP \cite{Group_aware_Contrastive_Regression_for_Action_Quality_Assessment}, C3D-AVG-MTL \cite{What_and_How_Well_You_Performed}, and TSA-Net \cite{TSA_Net_Tube_Self_Attention_Network_for_Action_Quality_Assessment}. We believe that the strengths of transformers can be leveraged in video understanding through a spatiotemporal self-attention mechanism, which could alleviate the limitations of CNNs.

While powerful, CNNs are designed to capture short-range spatiotemporal information; however, they are incapable of modeling dependencies that extend beyond their receptive field \cite{is_space_time_attention_all_you_need_for_video_understanding}. Stacking convolutional layers allows the network to extend the receptive field but it is still inherently limited in capturing long-term dependencies \cite{Going_Deeper_with_Convolutions, Very_Deep_Convolutional_Networks_for_Large_Scale_Image_Recognition, Exploring_Self_attention_for_Image_Recognition}. CNNs receptive field grows deeper as you go in the network; i.e. the first layers are unable to capture global dependencies. In contrast, the self-attention mechanism in transformers can be applied to capture long-range dependencies, which allows all layers in the network to attend to both local and global dependencies \cite{is_space_time_attention_all_you_need_for_video_understanding}.

Moreover, CNNs have a strong inductive bias regarding the spatial connectivity between the pixels and its translation invariance; this bias can limit the model’s expressivity when ample data is present. On the other hand, transformers impose a less restrictive inductive bias which renders them better suited for big-data \cite{On_the_Relationship_between_Self_Attention_and_Convolutional_Layers, Exploring_Self_attention_for_Image_Recognition}.

Inspired by these observations, we propose multiple transformer-based approaches to address AQA; specifically, we explore the usage of transformers in predicting the score of an Olympic dive. The end-goal of this work is to provide a foundation for transformer-based architectures to tackle AQA applications. Additionally, our secondary goal is to provide action quality scores to improve athletes’ performance and to eliminate any human error or bias component in judging diving competitions.

\section{Related Works}
\label{sec:related_works}

\noindent \textbf{Action Quality Assessment.} The problem of AQA deals with providing a quality score for a particular action by analyzing frames of videos. \citeauthor{What_and_How_Well_You_Performed} \cite{What_and_How_Well_You_Performed} introduced a new architecture for this task by implementing a multitask model -- C3D-AVG-MTL which involves learning spatiotemporal feature vectors to identify actions and provide quality scores along with caption generation for the videos. Furthermore, \citeauthor{What_and_How_Well_You_Performed} curated and released a new benchmark dataset, MTL-AQA, which has now become the standard dataset for conducting AQA on diving videos.
 
C3D-AVG-MTL utilizes the 3D ConvNet architecture (C3D) for feature extraction from videos. This architecture was introduced in \cite{Learning_spatiotemporal_features_with_3d_convolutional_networks} which was regarded as a success on such tasks due to its hierarchical representations of spatiotemporal data. These extracted video features are then passed to the regression modeling architecture to get the scores. The two-stream inflated 3D ConvNets (I3D) \cite{Quo_vadis_action_recognition_a_new_model_and_the_kinetics_dataset} builds on the 3D ConvNets by including an optical-flow stream component to the architecture which helps it provide better representations. This I3D architecture is used as part of \cite{Auto_Encoding_Score_Distribution_Regression_for_Action_Quality_Assessment} which also addresses the task of AQA on diving videos. It uses the I3D architecture for video feature extraction and proposes a new regression method Distribution Auto-Encoder (DAE) which maps the extracted video features to a score distribution. They assume the quality score to be a random variable and obtain the score by sampling the value from the score distribution.

\citeauthor{Group_aware_Contrastive_Regression_for_Action_Quality_Assessment} talk about the relations among videos which can provide important clues for more accurate action quality assessment during both training and inference \cite{Group_aware_Contrastive_Regression_for_Action_Quality_Assessment}. Specifically, the problem of AQA is reformulated as regressing the relative scores with reference to another video that has shared attributes (e.g., category and difficulty), instead of learning unreferenced scores. Following this formulation, a new Contrastive Regression (CoRe) framework is proposed to learn the relative scores by pairwise comparison, which highlights the differences between videos and guides the models to learn the key hints for assessment. In order to further exploit the relative information between two videos, the paper devises a group-aware regression tree to convert the conventional score regression into two easier sub-problems: coarse-to-fine classification and regression in small intervals.

\noindent \textbf{Transformers.} Transformer \cite{Attention_is_all_you_need} is a novel model architecture consisting of stacked encoder and decoder layers, which relies on a self-attention mechanism to derive long-range global dependencies between the input and output. Additionally, transformers were proposed to address the limitations of recurrent neural networks (RNN) in sequence modeling tasks. Also, the sequentiality in RNNs is a computational bottleneck and thus there was a need for a parallelizable architecture. Furthermore, for long sequences, RNN tends to forget the content of previous, distant positions which is addressed in transformers by accepting all previous inputs concurrently; this improvement allows transformers to capture long-range dependencies better. In \cite{Attention_is_all_you_need}, \citeauthor{Attention_is_all_you_need} emphasized the importance of utilizing multi-head self-attention mechanism; specifically, by dividing the input into several chunks and combining several self-attention heads, the model is able to attend to different segments of the input. Also, since transformers forgo the sequential nature in RNN, the elements in the input sequence retain their position by employing positional encodings to encapsulate the position of the element in a sequence.

\noindent \textbf{Transformers in computer vision.} Initially transformers were only applied to transduction model tasks such as language translation \cite{Attention_is_all_you_need} and other Natural Language Processing (NLP) related applications such as BERT \cite{Bert_Pretraining_of_deep_bidirectional_transformers_for_language_understanding} and GPT-2 \cite{Language_models_are_unsupervised_multitask_learners}; however, recent developments show that they’re applicable to a wider range of modalities and applications \cite{An_image_is_worth_16x16_words_Transformers_for_image_recognition_at_scale, Multiview_Transformers_for_Video_Recognition, Video_transformer_network, Training_Vision_Transformers_with_Only_2040_Images}. TSA-Net \cite{TSA_Net_Tube_Self_Attention_Network_for_Action_Quality_Assessment} introduced a single object tracker and proposed TSA module which is inspired by transformers that can efficiently generate rich spatiotemporal contextual information by adopting sparse feature interactions. TSA-Net is more efficient than non-local methods by at least 50\% less FLOPS while maintaining similar performance. TSA-Net predicts the quality of action per sub clip in the video. In the TSA mechanism, noise background information can be ignored to assess the quality of an action without affecting the outcome. In this way, the network pays more attention to the features containing athlete information, reducing the interference presented by irrelevant background information. TSA module consists of two steps: spatiotemporal tube generation, and tube self-attention operation. To test the effectiveness of TSA module, they used the network head capable of supporting multiple tasks, including classification, regression, and score distribution predictions. A change in the output size of the multilayer perceptron (MLP) block and the definition of the loss function can accomplish all tasks. The results demonstrate that TSA-Net can capture long-range contextual information and achieve high performance with minimal computing resources.

Based on the success of Transformer \cite{Attention_is_all_you_need} scaling in NLP, \citeauthor{An_image_is_worth_16x16_words_Transformers_for_image_recognition_at_scale} experiment with applying a standard Transformer directly to images, with the fewest modifications possible. This study focused on the direct application of Transformers to image recognition \cite{An_image_is_worth_16x16_words_Transformers_for_image_recognition_at_scale}. On multiple image recognition benchmark tests, Vision Transformer (ViT) approaches or beats state-of-the-art (SOTA) when pretrained on ImageNet-21k or JFT-300M datasets. A standard Transformer encoder is used to process an image as a sequence of patches. Coupled with pretraining on large datasets, this simple, yet ease of use, strategy works exceptionally well. With this knowledge, ViT can have similar or exceed the SOTA for most image classification datasets, while being relatively affordable to pretrain. Moreover, ViT can be extended by fine-tuning the pretrained model on a domain-specific dataset for other downstream tasks.

\citeauthor{is_space_time_attention_all_you_need_for_video_understanding} introduce TimeSformer, a novel convolution-free approach for video classification using a transformer encoder model that extends ViT \cite{is_space_time_attention_all_you_need_for_video_understanding}. Instead of naively utilizing joint space-time attention, the authors proposed divided space-time attention, which allows the query patch to attend to all of the patches in the same time step, as well as to attend to patches in the same spatial location in other timesteps. Coupled with a novel space-time positional embedding, TimeSformer was able to address video understanding applications with great success. Additionally, their architecture has the capability to train on longer videos, which previously was considered unfeasible.

Inspired by the recent advancements in transformers, for our proposed methods, we adapt transformers for AQA, specifically for videos, which has not been explored thoroughly yet. Our method leverages transformer's characteristics to effectively capture long-range dependencies, parallelize computation, and provide a wider receptive field for diving videos and predict a quality score.

\section{Approach}
\label{sec:approach}

In this section, we outline the proposed transformer-based architectures that aim to tackle AQA.

\subsection{Data Preprocessing}
\label{sec:data_preprocessing}

The MTL-AQA dataset contains video clips with different resolutions and lengths; therefore, it was essential for us to preprocess the frames to the same image dimensions of 224x224 pixels to match the input of the pretrained TimeSformer model \cite{is_space_time_attention_all_you_need_for_video_understanding}.

First, we employ standard normalization to simplify the data’s distribution; this was achieved by computing the mean and standard deviation of all frames in the training dataset, which was $(0.2719, 0.4617, 0.5961)$ and $(0.1870, 0.1881, 0.2604)$ respectively per RGB channel.
Second, we downsample the spatial size by scaling the short-side to 256 while maintaining the aspect ratio (16:9 for the dataset used). Finally, we utilized data augmentation techniques such as randomly cropping a region of 224x224 from the downsized video, as well as randomly horizontally flipping the images. Overall, these procedures were done to eliminate inconsistencies and ensure that each feature had similar value ranges.

\subsection{Frame Selection Methods}
\label{sec:frame_methods}

\begin{figure*}[!htb]
  \centering
  \subfloat[Random sampling]{\includegraphics[width=0.95\textwidth]{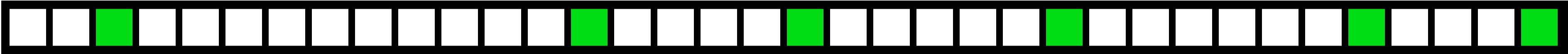}\label{fig:random_sampling}}
  \hfill
  \subfloat[Fixed-offset sampling]{\includegraphics[width=0.95\textwidth]{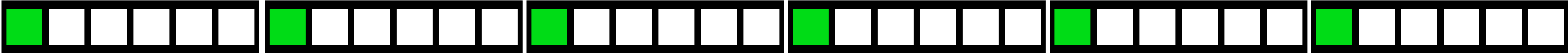}\label{fig:fixed_sampling}}
  \hfill
  \subfloat[Varied-offset sampling]{\includegraphics[width=0.95\textwidth]{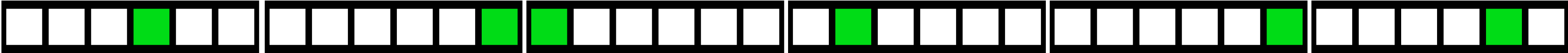}\label{fig:varied_sampling}}
  \hfill
  
  \caption{Visualization of different frame sampling methods, where green squares are the selected frames and the white squares are the skipped frames of a video clip. For the offset sampling methods, the entire clip is divided into $N$ equally-sized subclips.}
  \label{fig:sampling_methods}
\end{figure*}

The TimeSformer model \cite{is_space_time_attention_all_you_need_for_video_understanding} was pretrained on a specific number of frames, $N$, such as 8 or 96. Since the video clip’s duration could vary in length, we propose three frame selection methods listed below to ensure the model was provided the correct number of frames:

\begin{enumerate}[topsep=0pt,itemsep=-1ex,partopsep=1ex,parsep=1ex]
    \item \textbf{Random sampling}: Randomly sample $N$ frames from the video clip and then sort the sampled frames with respect to time, as shown in Fig. \ref{fig:random_sampling}.
    \item \textbf{Fixed-offset sampling}: Segment the video clip into $N$ non-overlapping subclips and select elements with a fixed-offset from each subclip, as shown in Fig. \ref{fig:fixed_sampling}.
    \item \textbf{Varied-offset sampling}: Segment the video clip into $N$ non-overlapping subclips and sample a random frame from each subclip, as shown in Fig. \ref{fig:varied_sampling}.
\end{enumerate}

Additionally, these methods also introduce a form of data augmentation as they will sample $N$ frames from the video clip in some manner and would yield different results on each run, which diversifies the training dataset and provides more data and variety to the transformer.

\subsection{MSE--Spearman Correlation Loss}
\label{sec:novel_loss}

As part of the preliminary experiments conducted, we observed that our models depicted no relationship between the model’s performance improving and the MSE loss decreasing. For instance, it was found that poorly performing models had similar MSE loss figures as models which performed significantly better. To counter this, we introduce a novel loss function which directly incorporates a differentiable Spearman correlation function \cite{Fast_Differentiable_Sorting_and_Ranking} as shown in Eqn \ref{eqn:novel_loss}.

\begin{equation}
    \textbf{L}(\textbf{y}, \hat{\textbf{y}}) = \alpha \cdot \textbf{MSE}(\textbf{y}, \hat{\textbf{y}}) - \beta \cdot \textbf{SpCorr}(\textbf{y}, \hat{\textbf{y}})
    \label{eqn:novel_loss}
\end{equation}

In Eqn \ref{eqn:novel_loss}, $\textbf{y}$ and $\hat{\textbf{y}}$ are the ground truth and predicted scores respectively. Additionally, the $\alpha$ and $\beta$ coefficients are responsible for assigning weights to the MSE loss and Spearman correlation respectively.

\subsection{Regression Target Variables}
\label{sec:regression}

The architectures proposed in this paper all aim to predict the same target values, the normalized score and the degree of difficulty, using the MSE–Spearman correlation loss as explained in Section \ref{sec:novel_loss}. The normalized score is a value from 0 to 1 that indicates the quality of the dive, which is computed by summing the judges’ votes and discarding the top and bottom two scores, followed by normalizing to $[0, 1]$ to conform to the diving competition scoring rules. A normalized score of 1 represents the best possible execution, while a score of 0 indicates a poorly executed dive. As for the degree of difficulty, it symbolizes the complexities of the performed dive, which is usually assigned by diving domain experts. Overall, the non-normalized final score of the dive would be the product of these two values which is computed by a post-processing step.

\subsection{I3D + MLP}
\label{sec:approach_i3d_mlp}

The I3D + MLP approach, which is the baseline model for our paper, is used by most existing works \cite{Auto_Encoding_Score_Distribution_Regression_for_Action_Quality_Assessment, TSA_Net_Tube_Self_Attention_Network_for_Action_Quality_Assessment, Group_aware_Contrastive_Regression_for_Action_Quality_Assessment}. In our implementation, the input video clips were encoded via a pre-trained I3D \cite{Quo_vadis_action_recognition_a_new_model_and_the_kinetics_dataset} model, followed by an MLP model with multiple dense layers to predict the score based on the features. The I3D model was pre-trained on the Kinetics-400 dataset to perform classification. Modifications were made to the I3D’s output layer; the original head layer is removed and is replaced by a linear layer that takes in an input of size 1024 and produces an output embedding of size 1024 which is later fed to an MLP block.

\subsection{I3D + Transformer Decoder}
\label{sec:approach_i3d_transformer_decoder}

Transformer Decoder \cite{Attention_is_all_you_need} has been introduced to the previous I3D + MLP model from Section \ref{sec:approach_i3d_mlp}. The latent embeddings from the I3D encoder are fed into the Transformer Decoder, whose output is then passed to an MLP block to get the score predictions.

\subsection{Transformer Encoder + MLP}
\label{sec:approach_transformer_encoder_mlp}

The Transformer Encoder + MLP model is an extension of the TimeSformer architecture \cite{is_space_time_attention_all_you_need_for_video_understanding} with a shallow MLP network as the decoder; the transformer encoder acts as the feature extractor whereas the MLP network attempts to decode the resulting encodings into an action quality score. The pretrained TimeSformer is initially trained for classification; nevertheless, due to the flexibility of transformers, we removed the head to extract the latent embedding representations which are then passed into the MLP decoder block.

\subsection{Transformer Encoder-Decoder}
\label{sec:approach_transformer_encoder_decoder}

Similar to Section \ref{sec:approach_transformer_encoder_mlp}, we extend the TimeSformer architecture \cite{is_space_time_attention_all_you_need_for_video_understanding} for feature extraction with a Transformer Decoder followed by an MLP head as the decoding component. This approach is an end-to-end model that exploits Transformer’s expressivity and recent advancements. The main motivation behind this architecture is the ability of Transformer Decoder to learn long-term dependencies better than an MLP as it uses self-attention mechanisms; additionally, Transformer Decoder can learn global dependencies better than an MLP because it uses a global attention mechanism \cite{Attention_is_all_you_need}.

\section{Experiments}
\label{sec:experiments}

To evaluate the proposed method, we carried out multiple experiments on the MTL-AQA \cite{What_and_How_Well_You_Performed} dataset. The results demonstrate that our proposed architectures outperform several previous SOTA models and achieve promising results as shown in Table \ref{tab:sota_comparison}. In the following sections, we elaborate on the experiments performed and analyze the prediction results provided by the models.

\subsection{Dataset}
\label{sec:experiments_dataset}

The dataset used in this paper was MTL-AQA \cite{What_and_How_Well_You_Performed} which is the largest dataset for AQA tasks, consisting of 1412 diving video clips. Each diving sample contains the detailed scoring of each judge as well as the diving difficulty degree. To abide by the evaluation protocol specified in the MTL-AQA original paper \cite{What_and_How_Well_You_Performed}, we utilized the same training and testing split of 1059 and 353 samples respectively to ensure comparable results between SOTA models.

\subsection{Evaluation Metric}
\label{sec:experiments_evaluation}

For the purpose of comparisons with previous works, we adopt Spearman correlation as the performance metric \cite{Assessing_the_Quality_of_Actions}. It measures the divergence between the ground truth score and the predicted score.

\begin{equation}
    \rho = \frac
    {\sum_{i} (p_i - p)(q_i - q)}
    {\sqrt{\sum_{i} (p_i - \bm\bar{p})^2(q_i - \bm\bar{q})^2}}
    \label{eqn:spearman_correlation}
\end{equation}

In Eqn. \ref{eqn:spearman_correlation}, $p$ and $q$ represent the ranking of the ground truth score and the predicted score respectively. A Spearman correlation of 0 indicates no correlation, whereas a value of 1 indicates perfect correlation.

\subsection{Implementation Details}
\label{sec:experiment_impl_details}

The models were trained using the MSE--Spearman correlation loss and implemented using the PyTorch framework. Two NVIDIA Tesla V100 GPUs were used to accelerate training along with eight CPUs of the AMD EPYC 7302 16-Core Processor @ 3.0GHz to accelerate data loading and processing. All experiments adopted AdamW optimizer \cite{Decoupled_Weight_Decay_Regularization} with the varied-offset sampling method for selecting frames from the video clips which are provided as the training data to the model. Overall, the models were trained over 200 epochs.

\subsubsection{I3D-based Models}
\label{sec:experiment_impl_i3d}

A pre-trained I3D model was used as the feature extractor in all experiments. With a batch size of 16 samples, a learning rate of 5e-5, and a weight decay of 1e-2. Furthermore, the I3D model's weights were not frozen while training. To evaluate diving action quality, the models were fine-tuned using the MTL-AQA dataset.

\noindent \textbf{I3D + MLP}. In order to predict action quality, the shallow MLP network with dropout was used instead of the MLP head of the I3D. This model used a dropout of 0.2 for all linear layers in the MLP block.

\noindent \textbf{I3D + Transformer Decoder}. The MLP head of the I3D was replaced with a Transformer Decoder followed by a shallow MLP block. The decoder takes the 1024 feature embeddings as the input and processes it using two sub-decoder layers and four heads. A shallow MLP block is then applied to the embeddings obtained from the decoder.

\subsubsection{Transformer-based Models}
\label{sec:experiment_impl_transformer}

For all the experiments, we utilized a pre-trained TimeSformer model \cite{is_space_time_attention_all_you_need_for_video_understanding} as the feature extractor. The batch size used was four samples, learning rate of 1e-5, and weight decay of 1e-5. Furthermore, we do not freeze the weights of the TimeSformer model while training the entire network. Moreover, the models were fine-tuned on the MTL-AQA dataset for diving action quality assessment. 

\noindent \textbf{Transformer Encoder + MLP}. The MLP head of the TimeSformer was replaced for regression to predict the action quality with a shallow MLP network with dropout regularization. The number of neurons in the shallow MLP was $(in=768, out=512)$, $(in=512, out=512)$, $(in=512, out=2)$ for the three layers respectively. This model used a dropout probability of 0.2 for all linear layers in the MLP block.

\noindent \textbf{Transformer Encoder–Decoder}. The MLP head of the TimeSformer was replaced with a Transformer Decoder which takes the 768 feature embeddings as the input and processes it using four sub-decoder layers with a dropout probability of 0.1. The resulting embeddings from the decoder are then passed to a final linear head.

\subsection{I3D + MLP Results}
\label{sec:exp_i3d_mlp}

For the I3D + MLP model, we experimented with 11 hyperparameters, where we found that three of the hyperparameters, namely, number of frames, batch size and learning rate, had the most significant impact and played a key role in how the results behaved with respect to these parameters. This baseline model achieves a Spearman correlation of 0.9226 trained over 450 epochs as shown in Table \ref{tab:sota_comparison}.

\subsubsection{Effects of Learning Rate}
\label{sec:exp_i3d_mlp_lr_epochs}

An important hyperparameter which influences the Spearman correlation score is the learning rate. Identifying the optimal learning rate is vital for establishing efficient learning. As shown in Table \ref{tab:i3d_mlp_batchsize}, we discerned that a learning rate of 5e-5 yields the highest Spearman  score of 0.9113. For lower learning rates, the Spearman correlation score plateaus since it converges towards the optima slowly which is prone to getting stuck in local optimas. Moreover, larger learning rates such as 1e-2 and 5e-3 are noisier and unstable in their gradient updates, which attains a Spearman correlation score of 0.4846 and 0.7497 respectively. This is majorly due to the model converging too quickly to a suboptimal solution.

\begin{table}[!htb]
    \centering
    \begin{tabular}{c|c}
        \toprule
        \textbf{Learning Rate} & \textbf{Sp. Corr.} \\
        \midrule
        1e-6 & 0.7064 \\
        \textbf{5e-5} & \textbf{0.9113} \\
        5e-4 & 0.9037 \\
        5e-3 & 0.7497 \\
        1e-2 & 0.4846 \\
        \bottomrule
    \end{tabular}
    \caption{Effect of learning rate on the I3D + MLP model's performance for 200 epochs.}
    \label{tab:i3d_mlp_batchsize}
\end{table}

\subsubsection{Effects of Number of Frames}
\label{sec:exp_i3d_mlp_numframes}

Regulating the number of frames in the I3D + MLP model is another impactful hyperparameter to consider. Each video has a varying number of frames and it is critical to find the optimal number of frames to maximize the extracted information. As per Table \ref{tab:i3d_mlp_numframes}, we noted that the best results were obtained when we used 64 frames with a Spearman correlation of 0.9226, whereas 16 frames obtained a score of 0.7990. Therefore, the higher the frame count, the better the results; this is as expected since more frames are being fed into the model to be analyzed and more information is extracted from the frames. Alternatively, the lower the frame count, the lesser the amount of information being fed  to the model resulting in a lower Spearman correlation score which is confirmed by the results.

\begin{table}[!htb]
    \centering
    \begin{tabular}{c|c}
        \toprule
        \textbf{\# Frames} & \textbf{Sp. Corr.} \\
        \midrule
        16 & 0.7990 \\
        32 & 0.9031 \\
        \textbf{64} & \textbf{0.9226} \\
        \bottomrule
    \end{tabular}
    \caption{Effect of number of frames on the I3D + MLP model's performance for 450 epochs.}
    \label{tab:i3d_mlp_numframes}
\end{table}

\subsection{I3D + Transformer Decoder Results}
\label{sec:exp_i3d_decoder}

The results in Table \ref{tab:sota_comparison} illustrates the I3D + Transformer Decoder Spearman correlation of 0.9317, which outperforms the I3D + MLP baseline of 0.9226.

\subsubsection{Effects of Batch Size}
\label{sec:exp_i3d_decoder_batch_size}

The batch size plays an essential role in how generalizable a model is. As we can observe in Table \ref{tab:i3d_decoder_batchsize}, the highest and lowest Spearman correlation, 0.9168 and 0.8898, were obtained via a batch size of 32 and 4 respectively. This signifies that a higher batch size is able to generalize better and attain more performant models since a larger batch size improves the convergence to a flatter minimizer. Consequently, a lower batch size is prone to sharp minimizers which degrades its generalizability as discussed in \cite{Training_set_batch_size}.

\begin{table}[!htb]
    \centering
    \begin{tabular}{c|c}
        \toprule
        \textbf{Batch Size} & \textbf{Sp. Corr.}\\
        \midrule
        4 & 0.8898 \\
        16 & 0.9105 \\
        \textbf{32} & \textbf{0.9168} \\
        \bottomrule
    \end{tabular}
    \caption{Effect of batch size on the I3D + Transformer Decoder model's performance for 450 epochs.}
    \label{tab:i3d_decoder_batchsize}
\end{table}

\subsubsection{Effects of Decoder Hyperparameters}
\label{sec:exp_i3d_decoder_lr_epochs}

The number of decoder layers and decoder heads play a crucial role in model performance. As per Table \ref{tab:i3d_decoder_numheadslayers}, we observed that using two decoder layers results in the highest Spearman correlation score. Increasing the number of layers causes the model to overfit on the data and degrades the performance of the model. Furthermore, when six or more decoder layers were used the Spearman correlation dropped to 0.5436. As for the decoder heads, we achieve comparable results using two or four heads, with four giving the best results of 0.9317. More than one decoder head is preferred since multiple heads can learn different patterns with each head. Further increasing the number of heads drastically increases the training time and number of parameters, but at the same time reduces the Spearman correlation due to overparameterization. 

\begin{table}[!htb]
    \centering
    \begin{tabular}{cc|c}
        \toprule
        \textbf{\# Heads} & \textbf{\# Layers} & \textbf{Sp. Corr.}\\
        \midrule
        \textbf{4} & \textbf{2} & \textbf{0.9317} \\
        4 & 6 & 0.9097 \\
        8 & 1 & 0.8587 \\
        8 & 6 & 0.5436 \\
        \bottomrule
    \end{tabular}
    \caption{Effect of number of heads and layers on the I3D + Transformer Decoder model's performance for 450 epochs.}
    \label{tab:i3d_decoder_numheadslayers}
\end{table}

\subsection{Transformer Encoder + MLP Results}
\label{sec:exp_encoder_mlp}

The results in Table \ref{tab:sota_comparison} show the Transformer Encoder + MLP predicted Spearman correlation of 0.9314 after 450 epochs, while the I3D + MLP baseline achieved 0.9226. 

\subsubsection{Effect of Different Frame-Sampling Methods}
\label{sec:exp_encoder_mlp_frame_sampling}

A significant hyperparameter is the choice of frame-sampling method used for the training of the network. As shown in Table \ref{tab:frame_sampling}, we performed comprehensive experiments by changing the sampling methods during the training for 200 epochs. We observed that the varied-offset sampling method provides the best Spearman correlation of 0.9284, whereas the fixed-offset sampling method provides the lowest Spearman correlation of 0.9218.

All of the aforementioned sampling methods are a form of data augmentation that add variety without losing the original data's semantic meaning. Each video clip yields a different subset of frames for each epoch, which synthetically expands the dataset; as discussed in \cite{Attention_is_all_you_need}, transformers are data hungry, where they scale extremely well the larger the dataset size becomes. However, fixed-offset imposes an unwarranted constraint that the frames must be spaced out evenly, which is empirically proven to be ineffective in contrast to random or varied-offset which loosens this constraint. In addition, random sampling is an aggressive data augmentation technique that may degrade the performance of the model as it could introduce high variance into the dataset that the model is incapable of capturing.

\begin{table}[!htb]
    \centering
    \begin{tabular}{c|c}
        \toprule
        \textbf{Frame Sampling Method} & \textbf{Sp. Corr.} \\
        \midrule
        Random & 0.9239 \\
        Fixed-offset & 0.9218 \\
        \textbf{Varied-offset} & \textbf{0.9284} \\
        \bottomrule
    \end{tabular}
    \caption{Effect of different frame sampling methods on the Transformer Encoder + MLP model's performance for 200 epochs.}
    \label{tab:frame_sampling}
\end{table}

\subsubsection{Effect of Pre-trained Models and Frame Count}
\label{sec:exp_encoder_mlp_pretrained}

TimeSformer \cite{is_space_time_attention_all_you_need_for_video_understanding} provides multiple model variants pretrained on Kinetics-400 \cite{k400}, Kinetics-600 (K600) \cite{k600}, Something-Something-V2 (SSv2) \cite{ssv2}, and HowTo100M \cite{howto100m} datasets. Each model has sub-variants based on the number of frames, spatial crop, video resolution, and duration of the videos trained on. We experimented with sampling a different number of frames (32, 48, and 64) for each of the pretrained models in Table \ref{tab:pretrained_effect}. Overall, we observed that the training duration and the frame count are positively correlated. Additionally, frame count and Spearman correlation are directly proportional; however, it is an interesting observation that the HowTo100M models do not adhere to this pattern. Our preliminary reasoning is that the K400 and K600 datasets are closely related to the MTL-AQA dataset due to their bias towards spatial scene information \cite{is_space_time_attention_all_you_need_for_video_understanding}; this allows the model to benefit greatly from a larger number of frames, but the HowTo100M dataset is dissimilar to the MTL-AQA dataset, which could indicate that the fine-tuning of the model with a higher number of sampled frames is degrading the performance as it is not able to utilize its pretrained embeddings. A deeper analysis of these results could be performed to confirm our understanding as part of our future work.

\begin{table}[!htb]
    \centering
    \begin{tabular}{cc|cc}
        \toprule
        \textbf{Model} & \textbf{\# Frames} & \textbf{Sp. Corr.} & \textbf{Train Period} \\
        \midrule
        HowTo100M & 32 & \textbf{0.9265} & \textbf{13h 19m} \\
        HowTo100M & 48 & 0.9218 & 18h 8m \\
        HowTo100M & 64 & 0.9149 & 24h 13m \\
        \midrule
        K400 & 32 & 0.9263 & \textbf{13h 16m} \\
        K400 & 48 & 0.9259 & 18h 33m \\
        K400 & 64 & \textbf{0.9280} & 23h 58m \\
        \midrule
        K600 & 32 & 0.9099 & \textbf{14h 45m} \\
        K600 & 48 & 0.9155 & 18h 25m \\
        K600 & 64 & \textbf{0.9246} & 25h 10m \\
        \bottomrule
    \end{tabular}
    \caption{Effect of pretrained model and frame count on the Transformer Encoder + MLP model's performance for 200 epochs.}
    \label{tab:pretrained_effect}
\end{table}

\subsubsection{Effect of Data Preprocessing}
\label{sec:exp_encoder_mlp_preprocessing}

As shown in Table \ref{tab:data_preprocessing}, we studied the effects of data preprocessing techniques on the performance of the Transformer Encoder + MLP model  as discussed in Section \ref{sec:data_preprocessing}. With no normalization or data augmentation, the model achieved a Spearman correlation of 0.9218. Additionally, as expected, when the data has been preprocessed with both normalization and data augmentation, we achieved a score of 0.9255.

Data augmentation expands the dataset’s size synthetically which helps with generalization and reduces overfitting \cite{The_Effectiveness_of_Data_Augmentation_in_Image_Classification_using_Deep_Learning}. As for normalization, it speeds up training and improves numerical stability by eliminating inconsistencies and ensuring that each feature has similar value ranges \cite{Verification_of_Normalization_Effects_Through_Comparison_of_CNN_Models}.

\begin{table}[!htb]
    \centering
    \begin{tabular}{l|c}
        \toprule
        \textbf{Method} & \textbf{Sp. Corr.} \\
        \midrule
        Transformer Encoder \& MLP & 0.9218 \\
         + Norm. \& Data Aug. & \textbf{0.9255} \\
        \bottomrule
    \end{tabular}
    \caption{Effect of normalization and data augmentation on the Transformer Encoder + MLP model's performance for 200 epochs.}
    \label{tab:data_preprocessing}
\end{table}

\subsubsection{Effect of MLP Topology}
\label{sec:exp_encoder_mlp_topology}

In the Transformer Encoder + MLP architecture, an important hyperparameter is the topology of the MLP. As shown in Table \ref{tab:topology_comp}, we experimented with MLP of varying depths and observed that we obtained the best Spearman correlation scores with a two-layer MLP topology with a score of 0.9288 whereas a single layer MLP achieved 0.9165.

The deeper the MLP block, the more parameters it contains; we posit that with a single layer, the model did not contain enough parameters to capture the data’s structure. Hence, expanding to two layers achieved the best result as it contained enough parameters to cover the dataset. Consequently, increasing beyond two layers overparameterized the model and incurred a slight penalty from overfitting on the dataset.

\begin{table}[!htb]
    \centering
    \begin{tabular}{c|c}
        \toprule
        \textbf{MLP Topology} & \textbf{Sp. Corr.} \\
        \midrule
        768 & 0.9165 \\
        \textbf{512 512} & \textbf{0.9288} \\
        768 768 & 0.9218 \\
        768 768 768 & 0.9244 \\
        768 768 768 768 & 0.9219 \\
        \bottomrule
    \end{tabular}
    \caption{Effect of the MLP topology on the Transformer Encoder + MLP model's performance for 200 epochs.}
    \label{tab:topology_comp}
\end{table}

\subsection{Transformer Encoder-Decoder Results}
\label{sec:exp_encoder_decoder}

The Transformer Encoder-Decoder model achieved a Spearman correlation of 0.9269 after 450 epochs, surpassing the baseline, as shown in Table \ref{tab:sota_comparison}.

\subsubsection{Effect of MSE--Spearman  Loss}
\label{sec:exp_encoder_decoder_loss}

To test the effectiveness of the loss, we analyzed its performance with the Transformer Encoder-Decoder architecture. As observed in Table \ref{tab:loss_comp}, the highest Spearman correlation score recorded, 0.9163, was for the architecture which gave equal weightage to the MSE loss and Spearman correlation. Similarly, the lowest obtained Spearman correlation score of 0.3429 is for the model which only uses the Spearman correlation as a metric to calculate the loss on. This leads us to believe that the Spearman correlation, by itself, is not a good loss metric because the differentiable approximation did not provide enough direction on how to robustly minimize the loss. Nevertheless, when used in conjunction with the MSE loss, it boosted the model’s performance by directly optimizing on the evaluation metric.

\begin{table}[!htb]
    \centering
    \begin{tabular}{cc|c}
        \toprule
        $\boldsymbol\alpha$ & $\boldsymbol\beta$ & \textbf{Sp. Corr.} \\
        \midrule
        0 & 1 & 0.3429 \\
        1 & 0 & 0.9050 \\
        1 & 10 & 0.9063 \\
        \textbf{1} & \textbf{1} & \textbf{0.9163} \\
        \bottomrule
    \end{tabular}
    \caption{Effect of MSE--Spearman correlation loss on the Transformer Encoder--Decoder model's performance for 200 epochs.}
    \label{tab:loss_comp}
\end{table}

\subsubsection{Effect of Weight Decay}
\label{sec:exp_encoder_decoder_weightdecay}

The weight decay hyperparameter was utilized as a regularization term to encourage smaller weights in the model which follows Ockham’s razor principle. With 0 and 1e-5 weight decay, the Spearman correlation was 0.9204 and 0.9142 respectively as illustrated in Table \ref{tab:weight_decay}. This was an interesting finding since weight decay acts as an imposed constraint on the model and lessening these constraints allowed the model to fully utilize its capabilities and degrees of freedom \cite{Decoupled_Weight_Decay_Regularization}. Overall, the excessive regularization was a hindrance to the model’s performance since regularization was already implemented in the form of dropout.

\begin{table}[!htb]
    \centering
    \begin{tabular}{c|c}
        \toprule
        \textbf{Weight Decay} & \textbf{Sp. Corr.} \\
        \midrule
        \textbf{0} & \textbf{0.9204} \\
        1e-5 & 0.9142 \\
        \bottomrule
    \end{tabular}
    \caption{Effect of weight decay on the Transformer Encoder--Decoder model's performance for 200 epochs.}
    \label{tab:weight_decay}
\end{table}

\subsection{Comparison with State-of-the-Art}
\label{sec:exp_sota}

\begin{figure*}[!htb]
  \centering
  \includegraphics[width=0.8\textwidth]{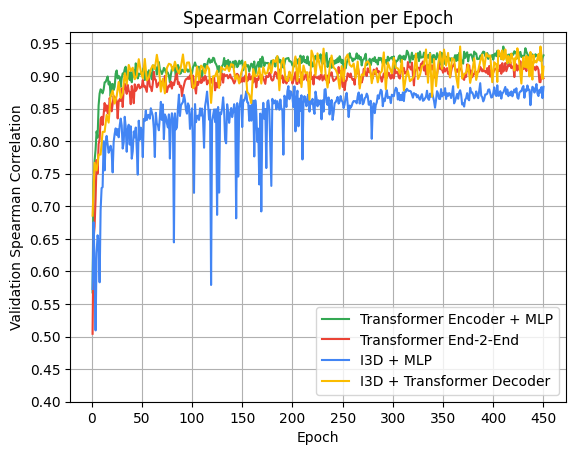}\label{fig:plot}
  \hfill
  \caption{The training progress and evaluation metric over 450 epochs for the proposed architectures.}
  \label{fig:model_comparison}
\end{figure*}

As shown in Figure \ref{fig:model_comparison}, the I3D + Transformer Decoder attained the best results  while the I3D + MLP produced the lowest results relative to the other proposed architectures. Even though the convolutional-free methods did not yield the highest results, they obtained comparable performance with convolutional-based methods, where 0.9314 and 0.9317 are the convolutional-free and transformer-convolution hybrid methods respectively. Overall, in comparison to the current SOTA as illustrated in Table \ref{tab:sota_comparison}, our proposed models obtain competitive Spearman correlation scores with an entirely new framework that is based on Transformers.

\begin{table}[!htb]
    \centering
    \begin{tabular}{c|c}
        \toprule
        \textbf{Model} & \textbf{Sp. Corr.} \\
        \midrule
        Pose + DCT \cite{Assessing_the_Quality_of_Actions} & 0.2682 \\
        ResNet 3D \cite{Learning_Spatio_Temporal_Features_with_3D_Residual_Networks_for_Action_Recognition} & 0.7633 \\
        MSCADC-MTL \cite{What_and_How_Well_You_Performed} & 0.8612 \\
        C3D-AVG-MTL \cite{What_and_How_Well_You_Performed} & 0.9044 \\
        MUSDL \cite{Uncertainty_aware_Score_Distribution_Learning_for_Action_Quality_Assessment} & 0.9273 \\
        TSA-Net \cite{TSA_Net_Tube_Self_Attention_Network_for_Action_Quality_Assessment} & 0.9393 \\
        NL-Net \cite{TSA_Net_Tube_Self_Attention_Network_for_Action_Quality_Assessment} & 0.9422 \\
        DAE-MT \cite{Auto_Encoding_Score_Distribution_Regression_for_Action_Quality_Assessment} & 0.9449 \\
        \textbf{CoRe + GART} \cite{Group_aware_Contrastive_Regression_for_Action_Quality_Assessment} & \textbf{0.9512} \\
        \midrule
        I3D + MLP & 0.9226 \\
        Transformer Encoder--Decoder & 0.9269 \\
        Transformer Encoder + MLP & 0.9314 \\
        \textbf{I3D + Transformer Decoder} & \textbf{0.9317} \\
        \bottomrule
    \end{tabular}
    \caption{Comparison with the state-of-the-arts on MTL-AQA.}
    \label{tab:sota_comparison}
\end{table}

\section{Conclusion}
\label{sec:conclusion}

In this report, we present multiple Transformer-based approaches for AQA along with our own implementation of the I3D + MLP baseline, which is able to capture rich spatiotemporal contextual information from videos. Overall, our proposed architectures achieve comparable results to the SOTA models, which magnifies the potential of using Transformers as a replacement for convolutional operators. This paper lays the foundation for Transformers in AQA applications and the possibility to extend to different video-based scenarios.

For future work, we plan to expand our computing resources to perform more experiments with larger batch sizes and number of epochs. Additionally, we plan to analyze the MSE–Spearman correlation loss with other applications such as video classification. Finally, we aim to evaluate our proposed methods on different datasets to assess its generalizability and performance.

\bibliography{acl2020}
\bibliographystyle{acl_natbib}

\end{document}